# Predicting Future Shanghai Stock Market Price using ANN in the Period 21-Sep-2016 to 11-Oct-2016

**Wanjawa, Barack Wamkaya**

School of Computing and Informatics,

University of Nairobi, Kenya

wanjawawb@students.uonbi.ac.ke

**ABSTRACT**

Predicting the prices of stocks at any stock market remains a quest for many investors and researchers. Those who trade at the stock market tend to use technical, fundamental or time series analysis in their predictions. These methods usually guide on trends and not the exact likely prices. It is for this reason that Artificial Intelligence systems, such as Artificial Neural Network, that is feedforward multi-layer perceptron with error backpropagation, can be used for such predictions. A difficulty in neural network application is the determination of suitable network parameters. A previous research by the author already determined the network parameters as 5:21:21:1 with 80% training data or 4-year of training data as a good enough model for stock prediction. This model has been put to the test in predicting selected Shanghai Stock Exchange stocks in the future period of 21-Sep-2016 to 11-Oct-2016, about one week after the publication of these predictions. The research aims at confirming that simple neural network systems can be quite powerful in typical stock market predictions.

# 1.0 INTRODUCTION

Stock markets deal in the business of buying and selling equity (stocks), though other trades in such markets also include bonds. Stocks tend to be the most traded instrument due to their general low value hence ease of entry and initial investment. Stock trade is therefore quite an active segment on any stock market. Stocks also tend to point the overall economic performance of a whole nation as the stock index which measures stock price movement is usually factored into the economic outlook of a particular country.

Buyers and sellers of stocks do not directly access the trading system. It is stockbrokers who do the trading on their behalf. Though the clients may decide on the offer prices, they usually rely on stockbroker advice, being the experts in the field. Getting the best deal out of the trade is the quest for any investor. Stockbrokers, or even investors themselves, would try to forecast or predict the best price of any stock for the date of trade or the future, in order to get maximum benefits (high returns, low expenditure). Factors to consider in forecasting include fundamentals of the particular company through technical analysis. Some even do time series analysis to get a pointer to the future prices. However, artificial intelligence (AI) can be used to develop predictive systems that are better in terms of showing the expected real prices, rather than just trends. Neural networks is the particular preferred AI technology that is used in developing predicting systems that can be applied for stock market predictions.

There are four widely recognized stock market prediction methods (NeuroAI, 2016). These are technical analysis (Huang et al., 2011, Deng et al., 2011), fundamental analysis (Chen et al., 2007), time series analysis (Zhang et al., 2008, Neto et al., 2009) and use of Artificial Intelligence (AI). Solving AI problems needs AI agents e.g. a learning agent such as Unsupervised learning, Reinforcement learning or Supervised learning (Cerna et al., 2005). Artificial Neural Networks (ANN) use supervised learning to train agents that can then be used for prediction.

ANN design is challenging, since it requires the choice or determination of optimum settings of various ANN aspects such as type of network (backpropagation, recurrent, feedforward), training method (unsupervised, reinforcement, supervised), data set partitioning/ratio for training and testing, number of nodes for input and output, number of hidden layers, size of hidden layers, number of training cycle repetitions, decision on activation function to use (threshold, linear, sigmoid, hyperbolic) and number of records to use for training/testing.

A previous research by the author (Wanjawa et al., 2014) already determined these ANN parameters and developed a model that was suitable for typical prediction as shown on the results presented in that research.

**Problem Statement:** ANNs have been tested with high success on next day predictions for stock markets. Is it possible to do actual future predictions, just relying only on what the ANN generates, without any knowledge from the market on the actual prices for stocks traded? Can this be tried for a typical stock exchange such as the Shanghai Stock Exchange?

## 2.0     USING ANN MODEL IN PREDICTION
### 2.1     ANN Model Configuration

The model used for prediction was that developed in an earlier research by the author (Wanjawa et al., 2014). The structure is shown in Fig. 1 below.

The model for next day stock price prediction is of configuration 5:21:21:1 i.e. 5 inputs, 2 hidden layers each with 21 neurons and finally, 1 output. The activation function used was sigmoid, with one bias for each layer. The data for training is from a 4-year prior period. The model was used to create a prototype using C# programming language. This program has an interface used for doing a training based on a loaded dataset in comma separated values (CSV) format. It also has a testing option based on a background CSV file that compares the predictions and the actual prices. This developed system was then used for prediction of stock prices at the Shanghai Stock Exchange (SSE).

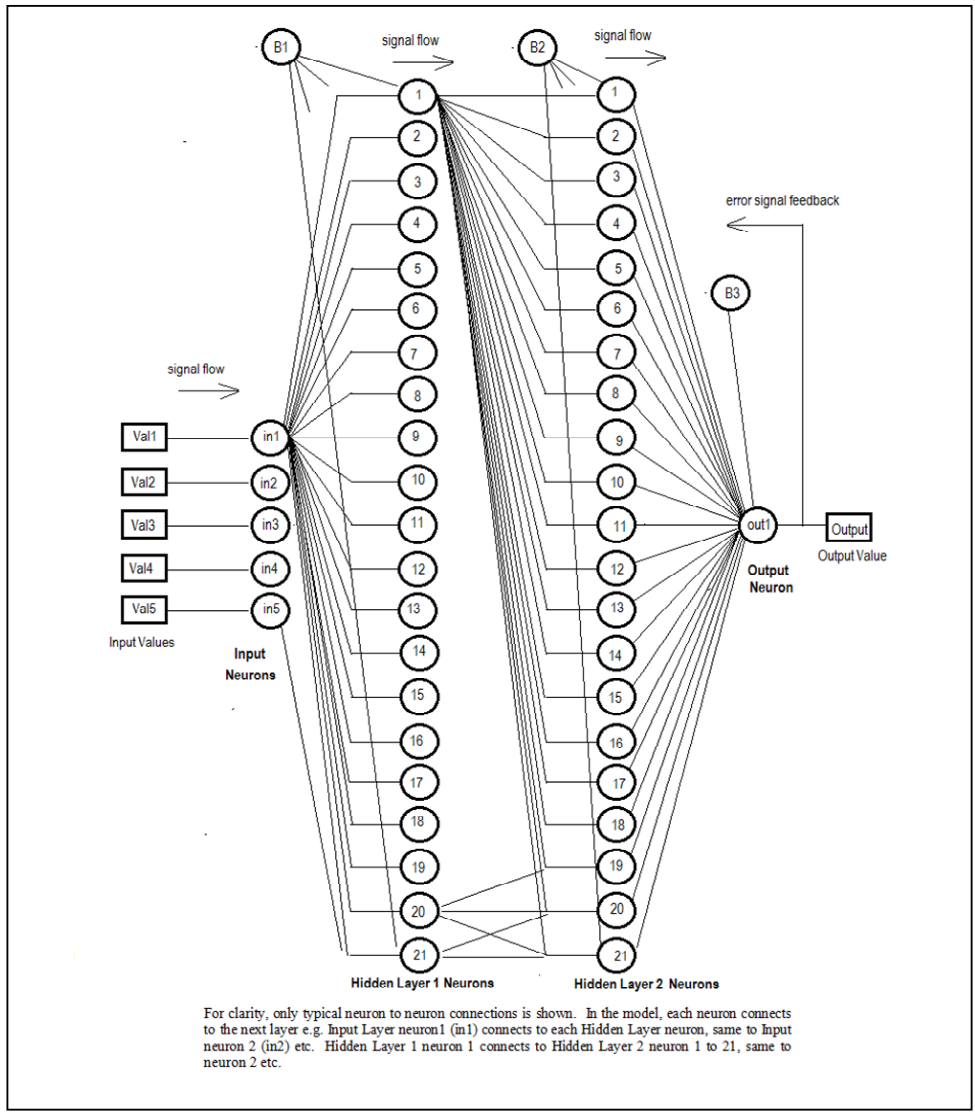

*Fig. 1* –ANN model For Stock Market Prediction (Source: Wanjawa et al., 2014)

**2.2     Selection of Stocks for Testing**

Seven final test stocks were selected from the Shanghai Stock Exchange (SSE) portfolio, out of an initial list of 12. The choice of stocks just consisted of the first ones on the numerical listing of the stocks, from 600000 to 600100. The selected stocks had also not been subjected to price adjustments. Price adjustments usually occur when there are company announcements that affect the otherwise normal price movement. Such announcements usually result into actions such as share splits or bonus issues.

The data sources usually specifically indicate these instances of price adjustments. The full list of stocks i.e. Domestic or A shares, as at September 2016 were 1,174, with reference numbers 600000 to 603999 (SSE, 2016).

The initial 12 stocks and the final 7 selected stocks are shown on the Table 1 below.

*Table 1* – *List of Selected Stocks from SSE (Source: SSE, 2016)*

| Code | Short name | Short name | Full name | Selected |
|---|---|---|---|---|
| 600009 | 上海机场 | SIA | Shanghai International Airport Co., Ltd. | N |
| 600010 | 包钢股份 | BSU | Inner Mongolia BaoTou Steel Union Co.,Ltd. | Y |
| 600015 | 华夏银行 | HUAXIA BANK | HUA XIA BANK CO., Limited | Y |
| 600016 | 民生银行 | CMBC | CHINA MINSHENG BANK | Y |
| 600028 | 中国石化 | Sinopec Corp. | China Petroleum and Chemical Corporation | Y |
| 600030 | 中信证券 | CITIC Securities Co., Ltd. | CITIC Securities Company Limited | N |
| 600031 | 三一重工 | SANY | SANY HEAVY INDUSTRY CO.,LTD | Y |
| 600035 | 楚天高速 | ChuTian Expwy | HUBEI CHUTIAN EXPRESSWAY CO.,LTD. | Y |
| 600048 | 保利地产 | PRE | POLY REAL ESTATE GROUP CO.,LTD | N |
| 600064 | 南京高科 | NJGK | NANJING GAOKE COMPANY LIMITED | N |
| 600089 | 特变电工 | TBEA | TBEA CO.,LTD. | Y |
| 600100 | 同方股份 | THTF | TSINGHUA TONGFANG CO., LTD | N |

Historical data for the seven selected stocks was then obtained for the period January 1, 2012 to September 14, 2016 for stocks 600010 (BSU, 2016), 600015 (HUAXIA BANK, 2016), 600016 (CMBC, 2016), 600028 (SINOPEC Corp., 2016), 600031 (SANY, 2016), 600064 (NJGK, 2016), 600089 (TBEA, 2016) and saved as separate CSV files. Out of this dataset, 4-year data i.e. January 1, 2012 to December 31, 2015 was used for training the prediction system, while 2016 data was used for testing.

## 2.3    Prediction Dates

The aim of the research was to predict stock prices for the selected 7 stocks in the 15-day period 21-Sep-2016 to 11-Oct-2016. The prediction was to be done without any knowledge of the actual day prices, since this was a prediction for a future period that is over 5 days from the current date.

## 2.4 Training and Testing Procedure

Seven separate networks were created using the configuration shown in section 2.1. Each network was then trained using the 4-year historical data (2012 - 2015) of the respective stock. For each network, five different tests were done, and the error obtained in the training recorded. Based on the error recordings, the best of the five test networks was retained as the final trained network for the particular stock, in readiness for prediction.

Testing was done per stock, by doing next day predictions for any date in the year 2016, upto September 14, 2016, just to confirm that the predicted values were close to the actual traded prices, based on the available actual trade price data. Figure 2 below shows the ANN prediction system with the Testing mode active.

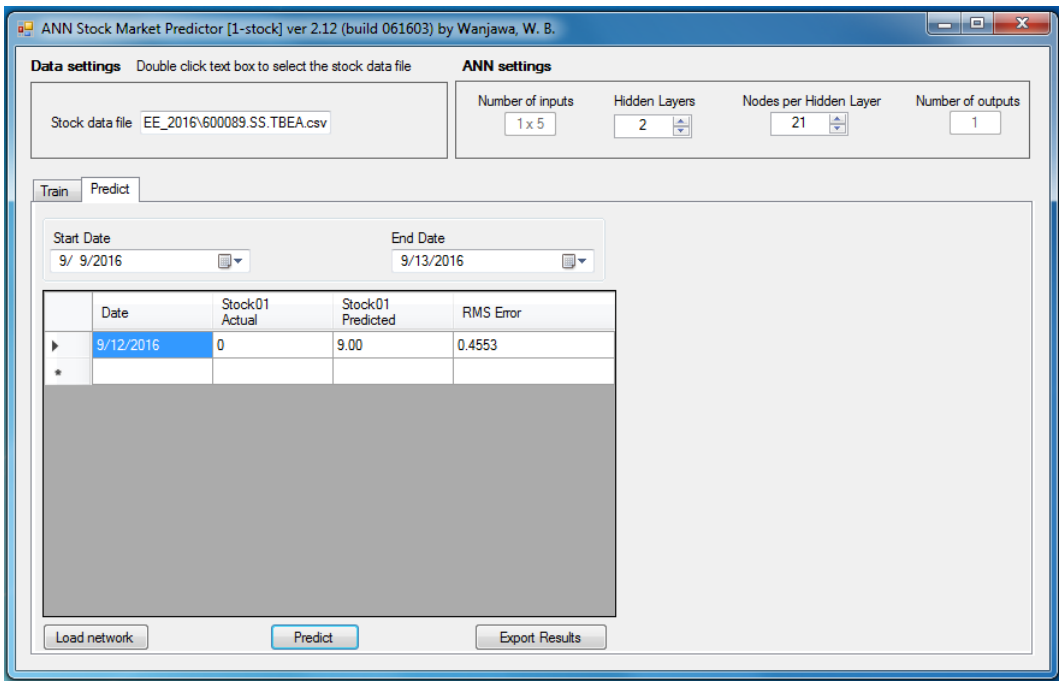

*Fig. 2 –ANN Prediction System – Predicting Mode (Source: Author)*

From September 15, 2016, there was no actual stock market price to compare the predictions with. It was therefore not possible to determine the error between actual stock price as should be published by the SSE, and the predicted price as generated by the prediction system. The CSV file with stock prices from Jan. 1, 2016 to Sep. 14, 2016 was extended to include working/trade days between Sep. 15, 2016 and Oct. 11, 2016 – a total of 19 days where prediction had to be done, based on no knowledge at all of the actual future prices of these days. The particular test period of Sep. 21 to Oct. 11, 2016 (15-days) was a response to the needs of an artist who needed predictions for these particular dates.

The final trained prediction network for the first SSE stock i.e. 600010 was fed with the CSV file described above, and instructed to predict the value for the next trade date after Sep. 14 i.e. to predict for Sep. 15. This was easy, since the system was reading the existing preceding 5 input values in order to predict for Sep. 15. The predicted value for Sep. 15 was then populated back to the CSV file. The model was then instructed to predict the value for Sep. 19. To do this, the model had to read the preceding 5 days' values, which were 4 actual price values plus a new predicted value for Sep. 15, in order to predict for Sep. 19. This new generated price was then populated back to the CSV file and the process of predict-populate continued until all the predictions had been obtained. The procedure was repeated for the second stock, until all predictions for the 7 selected stocks had been obtained. Being a next day prediction model, it was not possible to directly predict values from Sep. 21, unless the predictions started on the first unknown value i.e. Sep. 15, then predict all the way to Sep. 21, then continue on to Oct. 11, 2016. Note that the last 5 values are used to predict the next $6^{th}$ value.

## 3.0     RESULTS

The results obtained for prediction of each of the 7 stocks is shown in Table 2 below. There is no actual price to compare the predictions with, since these are all future values as at date of publication of research results. That means that the errors between predicted value and real traded value as published by the exchange cannot be calculated for now, in this research, until it comes to pass in the future.

*Table 2* – *Prediction for Seven Selected Stocks from Shanghai Stock Exchange*

| Stock | 600010 | 600015. | 600016 | 600028 | 600031 | 600064 | 600089 |
|---|---|---|---|---|---|---|---|
| ANN-model | T2 | T1 | T1 | T1 | T1 | T2 | T3 |
| Test-ID | T305224 | T108582 | T122736 | T120125 | T109318 | T62376 | T160457 |
| 15-Sep-16 | 2.73 | 10.70 | 9.02 | 4.93 | 5.33 | 16.73 | 8.95 |
| 16-Sep-16 | 2.69 | 10.31 | 8.97 | 4.93 | 5.56 | 16.58 | 8.94 |
| 19-Sep-16 | 2.69 | 9.63 | 9.03 | 4.93 | 5.49 | 16.48 | 8.94 |
| 20-Sep-16 | 2.69 | 9.84 | 9.14 | 4.93 | 5.28 | 16.35 | 8.91 |
| 21-Sep-16 | 2.70 | 10.03 | 9.07 | 4.93 | 5.08 | 16.20 | 8.91 |
| 22-Sep-16 | 2.71 | 10.48 | 8.89 | 4.93 | 5.17 | 16.08 | 8.89 |
| 23-Sep-16 | 2.72 | 10.33 | 8.70 | 4.93 | 5.23 | 15.98 | 8.89 |
| 26-Sep-16 | 2.72 | 10.55 | 8.63 | 4.93 | 5.40 | 15.87 | 8.89 |
| 27-Sep-16 | 2.72 | 10.48 | 8.69 | 4.93 | 5.38 | 15.78 | 8.88 |
| 28-Sep-16 | 2.71 | 10.03 | 8.80 | 4.93 | 5.28 | 15.72 | 8.88 |
| 29-Sep-16 | 2.71 | 10.25 | 8.79 | 4.93 | 5.10 | 15.67 | 8.87 |
| 30-Sep-16 | 2.70 | 10.62 | 8.69 | 4.93 | 5.15 | 15.63 | 8.87 |
| 03-Oct-16 | 2.71 | 10.09 | 8.59 | 4.93 | 5.21 | 15.61 | 8.87 |
| 04-Oct-16 | 2.70 | 10.22 | 8.49 | 4.93 | 5.35 | 15.61 | 8.87 |
| 05-Oct-16 | 2.71 | 9.79 | 8.48 | 4.93 | 5.43 | 15.62 | 8.87 |
| 06-Oct-16 | 2.70 | 9.89 | 8.52 | 4.93 | 5.36 | 15.64 | 8.87 |
| 07-Oct-16 | 2.71 | 10.02 | 8.53 | 4.93 | 5.24 | 15.68 | 8.87 |
| 10-Oct-16 | 2.70 | 10.17 | 8.49 | 4.93 | 5.06 | 15.73 | 8.87 |
| 11-Oct-16 | 2.71 | 10.36 | 8.42 | 4.93 | 5.08 | 15.80 | 8.87 |

It was noted that the predictions had some up and down movements for 5 of the stocks (600010, 600015, 600016, 600031 & 600064), while predictions for 2 others (600028 & 600089) tended to be just one constant value.

**4.0     CONCLUSION**

Artificial intelligence (AI) can be used for develop learning systems, such as Artificial Neural Network (ANN) prediction systems, which can then be used in typical prediction tasks, such as predicting stock market prices at typical stock exchanges. ANNs are powerful agents that use parallel computing as a basis of gaining intelligence from learning data, then use the gained knowledge in generating forecasts. A challenge in using ANNs has been the difficulty in formulating the model parameters. Using the model parameter of 5:21:21:1, with 4-year training data, as already determined in a previous research, an ANN prediction system was developed for next day prediction at a typical stock exchange.

Testing of ANN systems had tended to be based on predicting then comparing the prediction with the real data that is already known (testing data). In this particular research, the prediction was based on future values of stocks that are yet to be traded.

It was therefore a pure prediction into the future, with the accuracy of the prediction only possible when the actual future trade takes place.

This research was based on a fairly simple ANN configuration, which was only 2-layers deep, with each layer having only 21 neurons. This simplicity makes us believe that ANN configurations do not necessarily have to be complex to generate good prediction accuracy, as confirmed by next day prediction test results. We believe the same high accuracy should manifest when predicting real future values, where there is no reference at all to what the actual future prices shall be. This is the essence of this research paper.